\documentclass[]{spie}  

\usepackage{amsmath,amsfonts,amssymb}
\usepackage{graphicx}
\usepackage{tabularx}
\usepackage[colorlinks=true, allcolors=blue]{hyperref}

\title{Visual Search Patterns in 3D Pancreatic Imaging: An Eye Tracking Study}

\author[a]{Anna Anikina}
\author[a]{Leila Khaertdinova}
\author[b]{Trine Balschmidt}
\author[b]{Michael B Andersen}
\author[b]{Christoph F Müller}
\author[b]{Erik GS Brandt}
\author[b]{Henrik S Thomsen}
\author[e]{Claudia Mello-Thoms}
\author[a]{Bulat Ibragimov}
\affil[a]{Department of Computer Science, University of Copenhagen, Denmark}
\affil[b]{Department of Radiology, Herlev Hospital, Denmark}
\affil[e]{Department of Radiology, University of Iowa, United States}

\authorinfo{Send correspondence to Bulat Ibragimov:\\ E-mail: bulat@di.ku.dk}

\begin{document} 
\maketitle

\begin{abstract}
Eye tracking has emerged as a powerful tool for examining visual perception and search strategies in various domains, including medicine. While it is relatively straightforward to apply in 2D settings, its use in 3D medical imaging remains challenging and not yet well explored. This gap is particularly relevant for radiology, where volumetric images such as computed tomography (CT) scans are routinely read by medical experts. Radiologists typically interpret these images by navigating through hundreds of 2D slices, most often viewed in the axial projection. A taxonomy of eye movement data during navigation through a CT volume could be valuable to understand how radiologists approach diagnostic tasks. As an example of the derived taxonomy, we asked two radiologists to search abdominal CTs of the pancreas. We collect eye tracking data and align eye gaze movements with slice navigation to visualize the representation of the pancreas through volume and analyze clinicians' gaze behavior in both space and time. 
\end{abstract}


\section{INTRODUCTION}
\label{sec:intro}  

Analyses of visual search in 3D CT images present several complications compared to 2D images, primarily due to the added complexity of the depth (or slice) dimension \cite{drew2013scanners}. In 2D images, categorizing eye movements into fixations and saccades is a well-established technique for studying visual search behavior. However, this approach becomes problematic in 3D volumetric imaging \cite{drew2013scanners}. Studies have established that radiologists tend to adopt one of two dominant visual search strategies: “drilling” and “scanning”. Drillers restrict their eye movements to a relatively small region of the scan and then rapidly scroll through the stack of slices along the depth axis. Scanners, on the other hand, thoroughly search the entire x-y plane (the current 2D slice) before moving to the next slice in the stack \cite{drew2013scanners}. The studies dedicated to pulmonology found that drillers outperformed scanners in detecting lung nodules and covering lung tissue, suggesting that their strategy may be more effective for analyzing 3D CT images \cite{drew2013scanners,10.1117/1.JMI.3.1.015501}. 
In contrast, a study in digital breast tomosynthesis found no significant difference in diagnostic accuracy between scanners and drillers, additionally they found that breast radiologists follow a combination of “driller” and “scanner” strategies characterized by rapid, vigorous drilling with extensive scanning \cite{aizenman2017comparing}. 

There are still a very limited number of studies investigating visual search strategies in 3D volumetric medical images, and it remains unclear which approach—drilling, scanning, or potentially other as-yet-unidentified strategy—is most effective for different diagnostic tasks \cite{Wu2019}. We conducted a pilot study focused on visual search behavior during 3D CT analysis of the pancreas to help address these gaps. Our aim was to statistically characterize the strategies radiologists employ, to reveal new insights into the dynamics of volumetric search in pancreatic imaging, and to investigate whether their approaches align with existing classifications such as drilling and scanning, or if alternative strategies emerge in this specific diagnostic context.

\section{Dataset}

\subsection{Eye Tracking Experiment}

We asked two radiologists (Rad A and Rad B) to review CT images while wearing Pupil Labs Neon eye-tracking glasses \cite{Kassner:2014:POS:2638728.2641695}. They analyzed three axial CT images of locally advanced pancreatic cancer from the Pancreatic-CT-CBCT-SEG dataset \cite{Hong2021}, and three CT images of healthy patients from the Pancreas-CT dataset \cite{Roth2016}. All images were contrast-enhanced and had a resolution of 512×512 pixels. The images were visualized using the free IMAIOS DICOM Viewer \cite{Micheau2025}. The glasses, equipped with the calibration-free NeonNet gaze engine (approx. 1.3° accuracy), continuously recorded gaze behavior during the review sessions. Gaze data were then processed with the Dynamic RIM module from Pupil Labs to align eye-tracking with on-screen content \cite{pupil-labs_dynamic_rim_2025}.





\subsection{Preprocessing}

Pancreas segmentation was performed with TotalSegmentator \cite{Wasserthal2023}. We then aligned the original CT images, segmentation masks, and on-screen content using the ORB (Oriented FAST and Rotated BRIEF) feature detector \cite{6126544}, extracting and matching keypoints with a brute-force matcher and cross-checking for reliability. The matched keypoints enabled the estimation of a homography matrix to map the original CT image corners onto the on-screen scene. Because the segmentation masks correspond directly to the CT images, this provided a direct mapping of gaze data to the on-screen content.


   \begin{figure} [t]
   \begin{center}
   \begin{tabular}{c} 
   \includegraphics[width=0.97\textwidth]{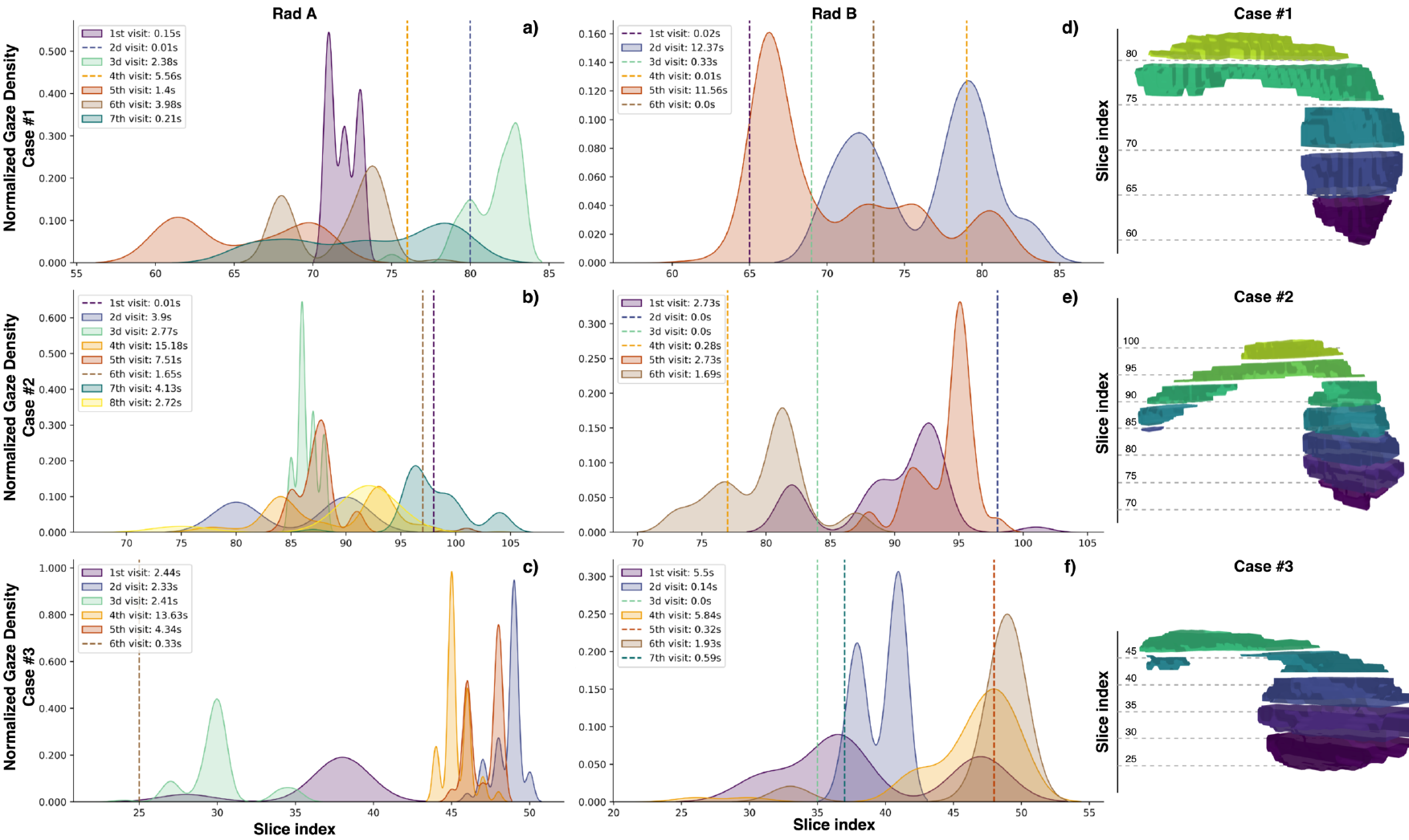}
   \end{tabular}
   \end{center}
   \caption[example] 
   { \label{fig:example2} Three example images analyzed by two radiologists are provided. Cases \#1, \#2, and \#3 show segmented pancreases with the slice index on the Y-axis. The same slice index numbers are repeated on the X-axis of the plots to indicate which slices the radiologists focused on during their reviews. Each plot (a-f) includes several curves representing the density of gaze data at specific time points, with the duration noted in the legend. The lines on the plots indicate the absence of variance in the gaze data for the current slice. 
}
   \end{figure} 

Each extracted gaze point represents the center of visual focus, but the human eye processes information across a wider area.
To approximate foveal vision -- visual angle $\theta = 1.5^\circ$ -- we calculated its equivalent size on the screen using a viewing distance of $D = 60$ cm and a screen resolution of $PPC = 38.4$ px/cm. The foveal radius is then defined by $r = \tan\left(\frac{\theta}{2}\right) \cdot D \cdot PPC$. This gives a foveal radius of $r = 30$ px in the original image, which was displayed within a 1000×700 px subregion of a 1920×1080 screen. When resizing the image to 512×512 px, we applied a uniform scaling factor $\alpha \sim 0.2667$, resulting in a final radius of $r \cdot \alpha = 8$ px to preserve the visual proportion of foveal coverage.

Radiologists analyze not only the pancreas but also the surrounding regions, which leads to multiple visits or returns (as shown in Fig. \ref{fig:example2}) to the pancreas during a single examination session. To quantify the durations of these focused visits, we divided timestamps into groups that represent separate visits to the pancreas region. We defined a jump threshold: $\theta = \mu_{\Delta t} + \alpha \cdot \sigma_{\Delta t}$, where $\Delta t$ is a difference between consecutive timestamps, $\mu_{\Delta t}$ is the mean of $\Delta t$ and $\sigma_{\Delta t}$ is a standard deviation. Any time gap larger than this threshold marks the start of a new group. 

\begin{table}[]
\caption{The table presents partial gaze statistics for two radiologists, both of whom analyzed the same first three cases. The columns labeled "mean", "median", "max", "std", and "total time" represent the time spent on the pancreas regions, with average, median, and variability in seconds across all visits, with "max" indicating the longest duration per visit. \textit{The approach we used for time separation is sensitive to the overall mean and standard deviation of time differences in the data; as a result, in some cases, similar time gaps may lead to slightly different outcomes}. "N" refers to the total number of switches to the pancreas region, while "Nr" denotes the number of revisits to the same slices.}
\label{tab:my-table}
\resizebox{\textwidth}{!}{%
\begin{tabular}{lccccccc}
\hline
\multicolumn{8}{l}{\textbf{Rad A:} \textit{Pancreas Gaze Only / Pancreas + Peripheral Gaze scenarios}} \\ \hline
Case & Mean (s)    & Median (s)  & Max (s)       & Std (s)     & Total Time (s) & N        & Nr       \\ \hline
\#1  & 1.96 / 3.35 & 1.40 / 1.59 & 5.56 / 13.69  & 1.99 / 4.35 & 13.69 / 26.83  & 40 / 71  & 8 / 39   \\
\#2  & 4.73 / 3.06 & 3.34 / 0.80 & 15.18 / 15.29 & 4.44 / 4.19 & 37.87 / 39.76  & 59 / 87  & 29 / 137 \\
\#3  & 4.25 / 1.35 & 2.43 / 0.50 & 13.63 / 11.34 & 4.35 / 2.45 & 25.48 / 32.39  & 76 / 175 & 14 / 58  \\ \hline
\multicolumn{8}{l}{\textbf{Rad B:}} \\ \hline
\#1  & 4.05 / 1.85 & 0.18 / 0.03 & 12.37 / 12.01 & 5.60 / 3.59 & 24.29 / 24.02  & 29 / 48  & 17 / 63  \\
\#2  & 1.24 / 2.03 & 0.99 / 1.17 & 2.73 / 6.98   & 1.20 / 2.27 & 7.43 / 20.32   & 25 / 43  & 15 / 60  \\
\#3  & 2.05 / 3.56 & 0.59 / 3.39 & 5.84 / 6.48   & 2.37 / 1.99 & 14.32 / 32.04  & 30 / 70  & 12 / 55  \\
\#4  & 2.97 / 1.81 & 2.48 / 1.08 & 6.92 / 5.16   & 2.43 / 1.81 & 14.86 / 18.13  & 18 / 42  & 18 / 84  \\
\#5  & 1.17 / 2.84 & 0.69 / 2.95 & 2.48 / 4.70   & 0.94 / 1.79 & 3.51 / 11.35   & 12 / 14  & 1 / 8    \\
\#6  & 4.31 / 3.62 & 1.68 / 2.04 & 14.05 / 16.51 & 5.22 / 5.13 & 25.85 / 28.98  & 26 / 38  & 42 / 69  \\ \hline
\end{tabular}%
}
\end{table}

\section{Results}

We collected statistics separately for two scenarios: (1) when the radiologists’ gaze was located strictly within the anatomical borders of the pancreas - referred to as the “Pancreas Gaze Only” scenario in Table \ref{tab:my-table} - and (2) when their gaze was directed in the vicinity of the pancreas, but the pancreas remained visible in their peripheral vision; for example, the gaze point $(x, y)$ lies outside the anatomical borders of the pancreas, but the pancreas is still included within the radius of foveal vision. The second scenario also includes all gaze instances from the first scenario, and we refer to it as “Pancreas + Peripheral Gaze” in Table \ref{tab:my-table}.


In scenario (1), Rad A had an average duration of $3.65$ seconds per visit across three cases, while Rad B averaged $2.44$ seconds per visit across three cases ($\#1-\#3$) and $2.63$ seconds across six cases ($\#1-\#6$). In scenario (2), the mean for Rad A dropped to $2.58$ seconds, as expected, whereas Rad B’s mean remained stable for both three and six cases. This pattern suggests that Rad B may prefer to focus on one organ at a time until their evaluation is complete, regardless of the broader anatomical context. Both radiologists exhibited large standard deviations within each case, indicating considerable variability in visit durations. To further understand this variability and the distribution of visit times, we calculated medians. When the mean and median are close, visit durations are more consistently distributed, suggesting that the radiologist spends a similar amount of time on most visits, for example, Rad A case $\#1$, Rad B cases $\#2, \#4, \#5$ in the (1) scenario, and Rad B cases $\#3, \#5$ in the (2) scenario. In contrast, when the mean is much higher than the median, for example, Rad B cases $\#1$ and $\#6$, this indicates many quick glances interspersed with occasional deep scrutiny, suggesting that certain areas prompted increased attention. Calculating the mean of medians for the (1), we found values of approximately $2.39$ seconds for Rad A and $1.09$ seconds for Rad B (six cases). In the (2), Rad A’s median dropped to approximately $0.96$ seconds, reflecting a more pronounced scanning strategy. In contrast, Rad B’s median slightly increased to $1.39$ seconds (six cases), consistent with the hypothesis that they maintain a one-organ-at-a-time approach and that their strategy remains stable across contexts. However, if we consider only three cases ($\#1-\#3$) for Rad B, the mean of medians increases from $0.58$ seconds in scenario (1) to $1.53$ seconds in scenario (2), which may indicate a possible shift toward a more thorough strategy. Another indicator supporting distinct reading strategies is the number of switches, or discrete visits, to the pancreas region. Although this measure varied substantially by radiologist, scenario, and case, Rad A consistently made more visits than Rad B. For the three cases ($\#1-\#3$) available for both readers, the mean number of switches for Rad A was $58$ compared to $28$ for Rad B in the (1), and $111$ vs. $53$ in the (2). This pattern again points to that Rad A employed a more scanning-oriented approach, revisiting the region frequently, whereas Rad B’s fewer entries may reflect a more targeted examination strategy.

While a one-organ-at-a-time approach might suggest a more thorough examination, our data do not support a direct relationship between this strategy and greater organ coverage. For example, in Cases $\#2$ and $\#3$, the percent coverage between the two radiologists was more similar, with Rad A sometimes covering more pancreas volume than Rad B (Case $\#2: 26.2\%$ vs. $21.4\%$ in (1), $34.1\%$ vs. $26.3\%$ in (2); Case $\#3: 32.2\%$ vs. $34.0\%$ in (1), $38.4\%$ vs. $40.7\%$ in (2)). Only in Case $\#1$, Rad B achieved much higher coverage than Rad A, with $30.3\%$ for Rad A versus $45.8\%$ for Rad B in scenario (1), and $36.2\%$ for Rad A versus $53.0\%$ for Rad B in scenario (2). Analysis of visit sequences revealed that both radiologists examined all major subregions (head, body, tail) of the pancreas, but no consistent pattern was observed in the order of region examination across cases.



\section{Conclusion}

In this study, we aimed to investigate the patterns and strategies that radiologists use when visually assessing the pancreas in CT images. Even with a small number of cases per reader, we observed some differences in gaze behavior and organ coverage, indicating the influence of individual search strategies. Our analysis suggests that one radiologist predominantly used a scanning approach, while the other adopted a different strategy that cannot be directly classified as scanning or drilling. However, due to the limited sample size, statistical comparisons between strategies were not feasible. Larger studies with more cases and participants are needed to robustly confirm and further explore these preliminary findings.

\acknowledgments 

This work was supported by the Novo Nordisk Foundation under Grant NFF20OC0062056, and the National Institutes of Health under grant 1R01CA259048. 

\bibliography{report} 
\bibliographystyle{spiebib} 

\end{document}